\documentclass[runningheads]{llncs}
\usepackage{graphicx}
\usepackage{microtype}
\usepackage{subfig}
\usepackage{booktabs} 
\usepackage{soul}
\usepackage{hyperref}
\usepackage{color,soul}
\usepackage{ctable}
\usepackage[bottom]{footmisc}
\newcommand*\samethanks[1][\value{footnote}]{\footnotemark[#1]}
\begin{document}
\title{COCO\_TS Dataset: Pixel--level Annotations Based on Weak Supervision for Scene Text Segmentation \thanks{This is a post-peer-review, pre-copyedit version of an article published in LNCS, volume 11730. The final authenticated version is available online at: \url{https://link.springer.com/chapter/10.1007/978-3-030-30490-4_26}}}
\titlerunning{COCO\_TS Dataset for Scene Text Segmentation}
%
\author{Simone Bonechi\thanks{Equal contribution} \and
Paolo Andreini\samethanks \and
Monica Bianchini \and
Franco Scarselli}

\authorrunning{S. Bonechi et al.}
%
\institute{DIISM, University of Siena, Via Roma 56, Siena, Italy}
\maketitle              
\begin{abstract}
The absence of large scale datasets with pixel--level supervisions is a significant obstacle for the training of deep convolutional networks for scene text segmentation. For this reason, synthetic data generation is normally employed to enlarge the training dataset. Nonetheless, synthetic data cannot reproduce the complexity and variability of natural images. In this paper, a weakly supervised learning approach is used to reduce the shift between training on real and synthetic data.
Pixel--level supervisions for a text detection dataset (i.e. where only bounding--box annotations are available) are generated. In particular, the COCO--Text--Segmentation (COCO\_TS) dataset, which provides pixel--level supervisions for the COCO--Text dataset, is created and released. 
The generated annotations are used to train a deep convolutional neural network for semantic segmentation. 
Experiments show that the proposed dataset can be used instead of synthetic data, allowing us to use only a fraction of the training samples and significantly improving the performances.

\end{abstract}

\section{Introduction}\label{Introduction}
Scene text segmentation is an important and challenging step in the extraction of textual information in natural images. It aims at making dense predictions in order to detect, for each pixel of an image, the presence of text. Convolutional Neural Networks (CNNs) are the state--of--the--art in many computer vision tasks, including scene text segmentation. Nonetheless, their training is usually based on large sets of fully supervised data. To the best of our knowledge, only two public datasets are available for scene text segmentation, i.e. ICDAR--2013 \cite{icdar2013} and Total--Text \cite{totalText}, that anyway contain a number of pixel--level annotated images barely sufficient to train a deep segmentation network. 
A solution to this problem has been proposed in \cite{tang2017scene}, where a pixel--level supervision is produced employing the synthetic image generator introduced by \cite{synthText}. 
However, unfortunately, there is no guarantee that a network trained on synthetic data will generalize to real images. This usually depends on the quality of the generated data (i.e. how much they are similar to real images), since the domain--shift may affect the generalization capability of the model. \\
\noindent
In this paper, we propose to employ weak supervisions to improve the performances on real data. Indeed, a lot of datasets for text localization, in which the supervision is given by bounding--boxes around the text, are available (f.i. COCO text \cite{cocotext}, ICDAR--2013 \cite{icdar2013}, ICDAR--2015\cite{icdar2015}, and MLT \cite{MLT}). In fact, obtaining this type of annotations is easier than providing a full pixel--level supervision, despite being less accurate. Inspired by \cite{bbox}, we adopt a training procedure that exploits these weak annotations. Specifically, the training procedure consists of two distinct steps.
 \begin{enumerate} 
\item A background--foreground network is trained on a large dataset of synthetically generated images with full pixel--level supervision. The purpose of this network is to recognize text within a bounding--box. 
\item A scene text segmentation network is trained on a text localization dataset, in which the pixel--level supervision is obtained exploiting the output of the background--foreground network.
 \end{enumerate} 
\noindent The logic behind this approach is that training a segmentation network focused on a bounding--box is a simpler task than using the entire image. In fact, inside a bounding--box, the text dimension is known (directly related to the box dimensions) and the background (i.e. non textual objects) variability is reduced. Moreover, the box annotation gives a precise information on the text position, since each pixel which is not included in a box does not represent text. Therefore, we exploit weak annotations to produce accurate pixel--level supervisions for a dataset of real images, which allows to reduce the domain--shift between synthetic and real data. In particular, employing the background--foreground network, the COCO--Text--Segmentation (COCO\_TS) dataset, which contains pixel--level segmentation supervisions for the COCO--Text dataset \cite{cocotext}, has been generated.
COCO\_TS will be made publicly available\footnote{\url{http://clem.diism.unisi.it/~coco_ts/}} to foster reproducibility and to promote future research in scene text segmentation. \\
\noindent
A series of experiments was conducted to evaluate the effectiveness of the proposed dataset compared to the use of synthetic data, as previously proposed in literature.
The obtained results suggest that, using the COCO\_TS dataset, a deep convolutional segmentation network can be trained more efficiently than using synthetic data, employing only a fraction of the learning set. Moreover, it is worth noting that the proposed procedure, used for COCO--Text, can be applied to generate pixel--level supervisions for any text localization dataset annotated at the bounding--box level. \\
\noindent
The paper is organized as follows. In Section \ref{Related_Works} related works are briefly reviewed. In Section \ref{method} the COCO\_TS generation procedure is described. Section \ref{Experiments} reports the experimental setup and the results obtained in scene text segmentation on the ICDAR--2013 and Total--Text datasets. Finally, some conclusions are drawn in Section \ref{Conclusions}.

\section{Related Works} \label{Related_Works}
The proposed method is related to four main research topics, namely synthetic data generation, bounding--boxes for semantic segmentation, semantic segmentation with CNNs, and scene text segmentation, whose literature is reviewed in the following.

\paragraph{Synthetic data generation.} 
Synthetic datasets are a cheap and scalable alternative to the human ground--truth supervision in machine learning. Recently, several papers reported on the use of synthetic data to face a variety of different problems. Large collections of synthetic images of driving scenes in urban environments were generated in \cite{SYNTHIA}, synthetic indoor scenes have been exploited by \cite{indoor}, while artificial images of Petri plates were created in \cite{synth_colony}. In text analysis, \cite{jaderberg_text} proposed the use of synthetic data for text spotting, localization and recognition. An improved synthetic data generator for text localization in natural images was proposed by \cite{synthText}. This synthetic data generator engine has been modified in \cite{tang2017scene} to extract pixel--wise segmentation annotations. Similarly to \cite{tang2017scene}, in this work, the engine proposed in \cite{synthText} was used for scene text segmentation.

\paragraph{Bounding--boxes for semantic segmentation.} 
In order to reduce the data labeling efforts, weakly supervised approaches aim at learning from weak annotations, such as image--level tags, partial labels, bounding--boxes, etc. Bounding--box supervision was used to aid semantic segmentation in \cite{BoxSup}, where the core idea is that to iterate between automatically generating region proposals and training convolutional networks. Similarly, in \cite{EM_weak}, an Expectation--Maximization algorithm was used to iteratively update the training supervision. Instead, in \cite{bbox_khoreva}, a GrabCut--like algorithms is employed to generate training labels from bounding boxes. Finally, more related to this work, in \cite{bbox}, the segmentation supervision for a semantic segmentation network is directly produced from bounding--box annotations, exploiting a deep CNN.

\paragraph{Semantic segmentation with CNNs.} 
Image semantic segmentation aims at inferring the class of each pixel of an image. Recent semantic segmentation algorithms often convert existing CNN architectures, designed for image classification, to fully convolutional networks \cite{FCN}. These networks have generally an encoder--decoder structure. Moreover, the level of details required by semantic segmentation inspired the use of dilated convolution to enlarge the receptive field without decreasing the resolution \cite{atrous}.  
Besides, different solutions have been proposed to deal with the presence of objects at different scales. The Pyramid Scene Parsing Network (PSPNet) \cite{PSP} applies a pyramid of pooling to collect contextual information at different scales. Instead, Deeplab \cite{deeplab} employs atrous spatial pyramid pooling, which consists of parallel dilated convolutions with different rates.

\paragraph{Scene Text Segmentation.} 
Document image segmentation has a long history and was originally based on thresholding approaches (local, global or adaptive) \cite{otsu,binarization,laplacian}. The application of these methods to scene text segmentation is quite challenging, due to the high variability of conditions that can be found in natural images. To face this variability, in \cite{bai2014seed}, low level features are used to identify the seed points of texts and backgrounds and then to segment the text using semi--supervised learning. In \cite{mishra2011mrf}, the binarization of scene text has been formulated as a Markov Random Field model optimization problem, where the optimal binarization is obtained iteratively with Graph Cuts. To improve the segmentation performance, a multilevel maximally stable extremal region approach, applied together with a text candidate selection algorithm based on hand--extracted text--specific features, has been presented in \cite{tian2014scene}. Finally, in \cite{tang2017scene}, a CNN approach to scene text segmentation is described, which employs three stages for extraction, refinement and classification.

\section{Materials and Methods}\label{method}
In the following, a general overview of the proposed method is provided. The sets of data involved in the creation of the COCO\_TS dataset are introduced in Section \ref{Benchmark Datasets}. Section \ref{COCO_Text_Segmentation} describes the weakly supervised approach used to generate COCO\_TS and finally, in Section \ref{Scene_Text_Segmentation}, the COCO\_TS dataset is used to train a deep segmentation network.

\subsection{Datasets} \label{Benchmark Datasets}

\paragraph{Synthetic dataset.} 
In this work, the same generation process proposed by \cite{synthText} has been employed to create a large set of synthetic scene text images. The engine renders synthetic text to existing background images, accounting for the local three dimensional scene geometry. A synthetic dataset of about 800000 images was generated following this procedure. 
From this set of images, about 1000000 image crops have been extracted. Specifically, for each word, a bounding--box is defined and enlarged by a factor of 0.3, and then the image is cropped around the bounding--box. These bounding boxes have been used to train the background--foreground network described in Section \ref{sup_generation}.

\paragraph {COCO--Text.}
The COCO--2014 dataset \cite{coco}, firstly released by Microsoft Corporation, collects instance--level fully annotated images of natural scenes. COCO--Text \cite{cocotext} is based on COCO--2014 and contains a total of 63686 images, split in 43686 training, 10000 validation, 10000 test images, supervised at the bounding--box level for text localization. Differently from other scene text datasets, the COCO--2014 dataset was not collected specifically for the extraction of textual information, hence some of its images do not contain text. Therefore, for the generation of the proposed COCO\_TS dataset, a subset of 14690 images have been selected from COCO--Text, each one at least including a bounding--box labeled as legible, machine printed, and written in English.

\paragraph {ICDAR--2013.}
The ICDAR--2013 \cite{icdar2013} dataset collects a training and a test set containing 229 and 233 images, respectively. The images are extracted from ICDAR--2011 \cite{icdar2011}, after the removal of duplicated images and with some revisited ground--truth annotations. The scene text segmentation challenge in the ICDAR--2015 competition \cite{icdar2015} was based on the same datasets as ICDAR--2013.

\paragraph {Total--Text.}
Total--Text \cite{totalText} is a scene text dataset which collects 1255 training images and 300 test images with a pixel--level supervision. Differently from ICDAR--2013, where texts have always a horizontal appearance, this dataset contains images with texts showing highly diversified orientations.

\subsection{COCO\_TS Dataset}\label{COCO_Text_Segmentation}
\begin{figure*}[ht]
\begin{center}
\centerline{\includegraphics[scale=0.42]{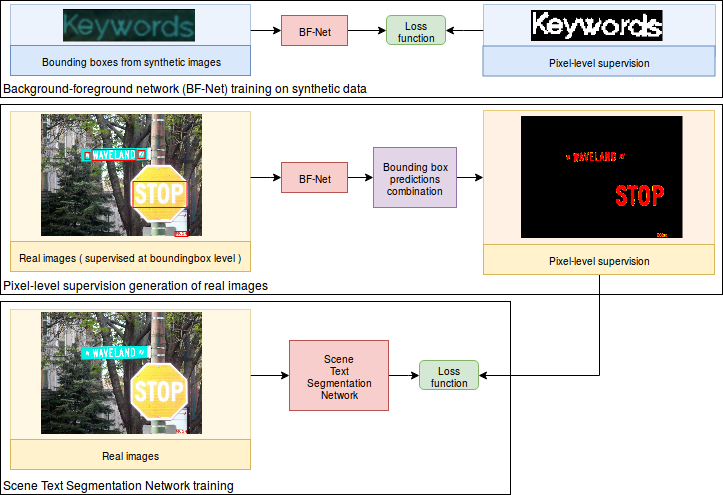}}
\caption{Scheme of the supervision generation procedure. Steps 1. to 3. are sketched starting from the top.}
\label{training_scheme}
\end{center}
\end{figure*}

Collecting supervised images for scene text segmentation is costly and time consuming. In fact, only few datasets with a reduced number of images are available. Instead, numerous datasets provide bounding--box level annotations for text detection. In this paper, we introduce the COCO\_TS dataset, which provides 14690 pixel--level supervisions for the COCO--Text images. The supervision is obtained from the available bounding--boxes of the COCO--Text dataset exploiting a weakly supervised algorithm. The supervision generation procedure is explained in the following and summarized in Figure \ref{training_scheme}.
\subsubsection{Supervision Generation of COCO\_TS \hspace{6cm}}\label{sup_generation}

The supervision generation consists of three different steps.
 \begin{enumerate} 
\item A background--foreground network is trained on synthetic data to extract text from bounding--boxes. 
\item The background--foreground network is employed to generate pixel--level supervisions for real images of the COCO\_TS dataset. 
\item A scene text segmentation network is trained on the real images with the generated supervisions.
 \end{enumerate} 
\paragraph{Background--Foreground Network.}
A deep neural network is trained to segment the text inside a bounding--box, thus separating the background from the foreground. The rationale beneath the proposed approach is that realizing a background--foreground segmentation, constrained to a bounding--box, is significantly simpler than producing the segmentation of the whole image. For this reason, we suppose that even if trained on synthetic data, the background--foreground network can effectively be used to segment text in bounding boxes extracted from real images. To train the background--foreground network, pixel--level supervisions of a significant number of bounding--boxes is required. The 1000000 bounding--box crops extracted from the synthetic dataset have been used to this purpose. 

\begin{figure*}[ht]
\begin{center}
\centerline{
\includegraphics[height=3.2cm,keepaspectratio]{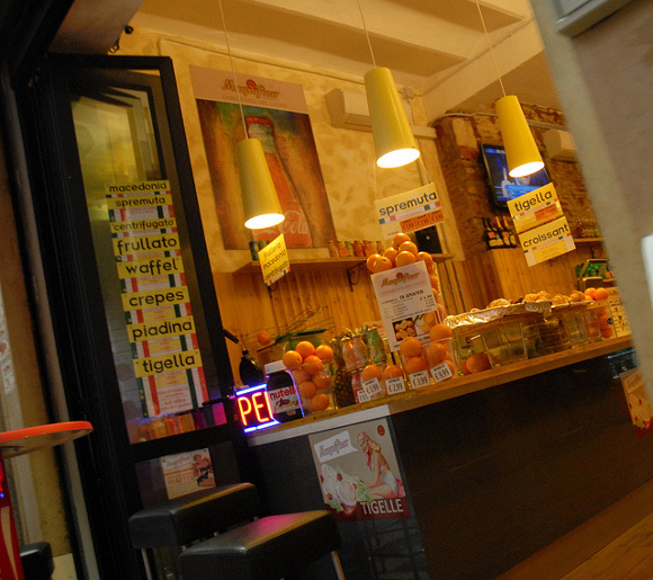}
\includegraphics[height=3.2cm,keepaspectratio]{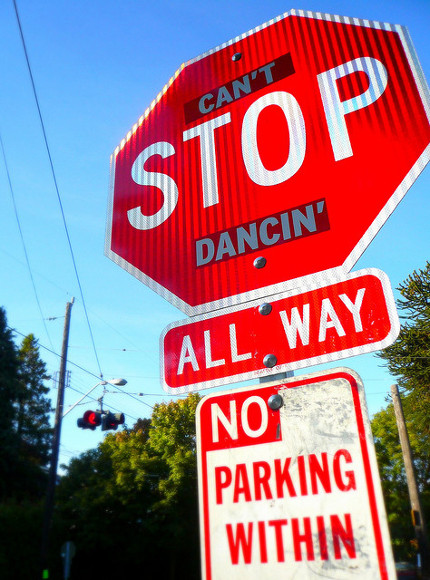}
\includegraphics[height=3.2cm,keepaspectratio]{generated_supervision/COCO_train2014_000000014886.jpg}}
\vskip 0.08in
\centerline{
\includegraphics[height=3.2cm,keepaspectratio]{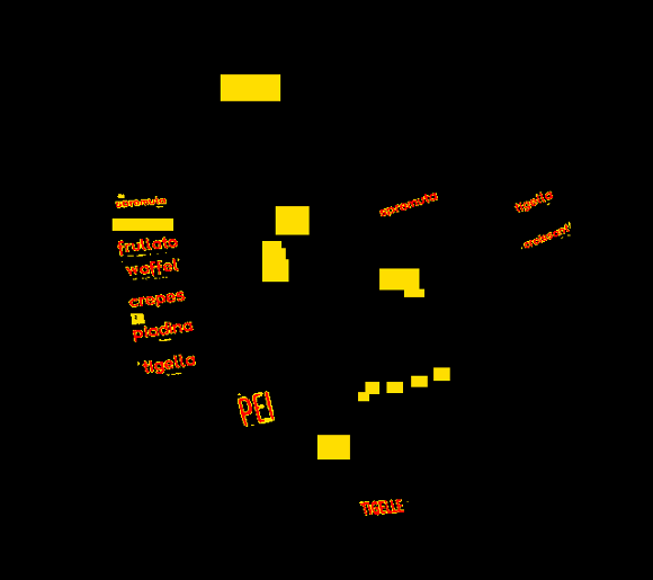}
\includegraphics[height=3.2cm,keepaspectratio]{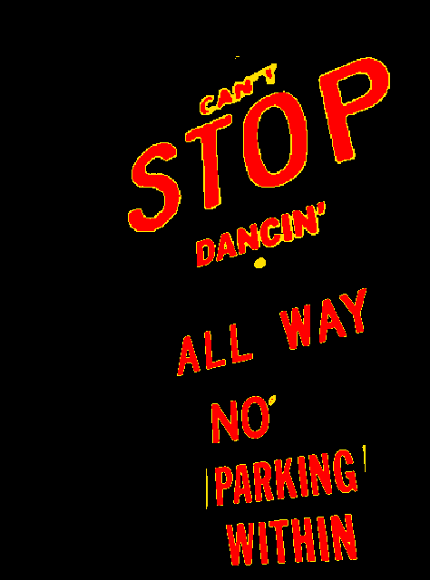}
\includegraphics[height=3.2cm,keepaspectratio]{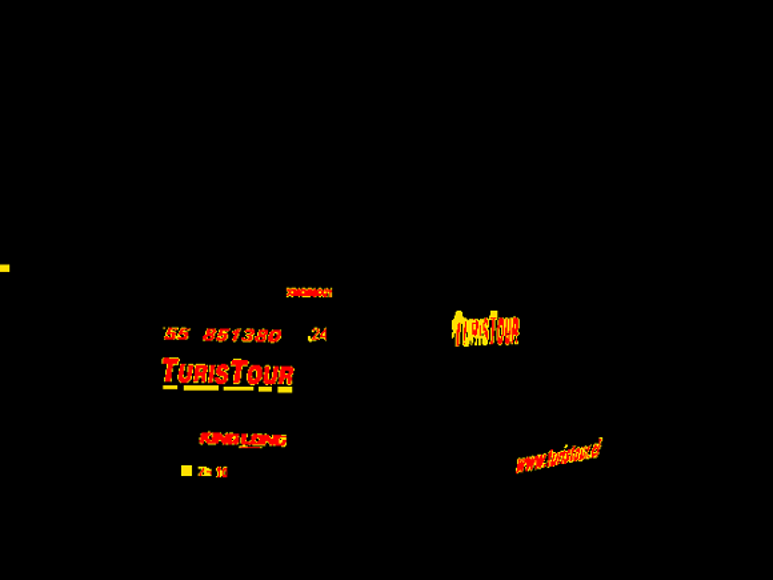}}
\caption{The original images and the generated supervisions, on the top and at the bottom, respectively. The background is colored in black, the foreground in red, and the uncertainty region in yellow.}
\label{supervision_example}
\end{center}
\end{figure*}

\paragraph{Pixel--level supervision generation.}
After the training phase, the background--foreground network is applied on the bounding--boxes extracted from the COCO--Text dataset.
For each image, the pixel--level supervision is obtained combining the probability maps (calculated by the background--foreground network) for all the bounding--boxes inside the image. In those regions where bounding--box annotations overlap, the prediction with the highest foreground probability value is considered. The final pixel--wise annotation $l(x, y)$, at position $(x, y)$, is obtained employing two fixed thresholds, $th_1$ and $th_2$, on the probability maps $prob(x,y)$:
  \begin{equation}
    l(x, y)=\left\{
                \begin{array}{ll}
                 background \quad \quad \textrm{if} \quad prob(x, y) < th_1 \\
                 foreground \quad \quad \textrm{if} \quad prob(x, y) > th_2\\
                  uncertain \quad \quad \quad \hspace*{-1mm}  \textrm{otherwise}
                \end{array}
              \right.
\label{FixedTh}
\end{equation}
The two thresholds $th_1$ and $th_2$ have been fixed to $0.3$ and $0.7$, respectively. 
If $prob(x, y) \in (th_1,th_2)$, then $(x,y)$ is labeled as uncertain. To provide a significant pixel--level supervision, bounding--boxes that are not labeled as legible, machine printed and written in English have been added to the uncertainty region. This procedure has been used to extract the COCO\_TS dataset.
Some examples of the obtained supervisions are reported in Figure \ref{supervision_example}.

\subsection{Scene Text Segmentation} \label{Scene_Text_Segmentation}
The COCO\_TS dataset is used to train a deep segmentation network (bottom of Figure \ref{training_scheme}) for scene text segmentation of both the ICDAR--2013 and Total--Text datasets. The effects obtained by the use of the COCO\_TS dataset, as an alternative to synthetic data, will be described in the next section.

\section{Experiments} \label{Experiments}
In the following, our experimental setup is shown. In particular, Section \ref{PSP} and Section \ref{training_details} introduce the segmentation network and define the implementation details used in our experimental setup. In Section \ref{coco_eval}, the generated annotations for the COCO\_TS dataset are evaluated, whereas Section \ref{sts_eval} assesses the insertion of the COCO\_TS dataset during the training of a scene text segmentation network.

\subsection{PSPNet}\label{PSP}
All the experiments are carried out with the PSPNet architecture \cite{PSP}, originally designed for semantic segmentation of natural images. This model, like most of the other semantic segmentation networks, takes an image as input and produces a per--pixel prediction. The PSPNet is a deep convolutional neural network, built on the ResNet model for image classification. To enlarge the receptive field of the neural network, a set of dilated convolutions replaces standard convolutions in the ResNet part of the network. The ResNet encoder produces a set of feature maps and a pyramid pooling module is used to gather context information. Finally, an upsample layer transforms, by bilinear interpolation, the low--dimension feature maps to the resolution of the original image. A convolutional layer produces the final per--pixel prediction. In this work, to better handle the presence of thin text and similarly to \cite{tang2017scene}, we modified the network structure adding a two level convolutional decoder.

\subsection{Implementation Details}\label{training_details}
The PSPNet architectures, used both for the background--foreground network and for scene text segmentation, are implemented in TensorFlow. Due to computational issues, in this work, the PSPNet based on the ResNet50 model is used as the CNN encoder. The experiments are realized based on the training procedure explained in the following.
As far as the background--foreground network is considered, the image crops are resized so that the min side dimension is equal to 185, while maintaining the original aspect--ratio. Random crops of $185\times185$ are used during training. Instead, for the scene text segmentation network, the input images have not been resized, and random crops of $281\times281$ are extracted for training. A multi--scale approach is employed during training and test. In the evaluation phase, a sliding window strategy is used for both the networks. The Adam optimizer \cite{adam}, with a learning rate of $10^{-4}$, has been used to train the network. The experimentation was carried out in a Debian environment, with a single NVIDIA GeForce GTX 1080 Ti GPU.

\subsection{Evaluation of the Supervision Generation Procedure}
\label{coco_eval}
The quality of the generation procedure cannot be assessed on COCO--Text, 
due to the absence of pixel--level targets. Therefore, we used the ICDAR--2013 dataset for which ground--truth labels are available.
Following the procedure described in Section \ref{sup_generation}, the segmentation annotations for the ICDAR--2013 test set have been extracted and compared to the original ground--truth. The results, measured using the pixel--level precision, recall and F1 score, are reported in Table \ref{annotation_res}. For this analysis, the uncertainty region has been considered as text. 
\begin{table*}[ht]
\begin{center}
\begin{small}
\begin{sc}
\begin{tabular}{lccc}
\toprule
 & Precision & Recall & F1 Score \\
\midrule
Proposed approach & 89.10\% & 70.74\% & 78.87\% \\
\midrule
\bottomrule
\end{tabular}
\end{sc}
\end{small}
\end{center}
\caption{Results of the annotation generation approach on the ICDAR--2013 test set.}
\label{annotation_res}
\end{table*}
\noindent
A qualitative evaluation of the generated supervision for the COCO\_TS dataset is reported in Figure \ref{supervision_example}.

\subsection{Scene Text Segmentation evaluation}
\label{sts_eval}
\begin{figure}[!ht]
\begin{center}
\centerline{
\includegraphics[width=2.25cm,keepaspectratio]{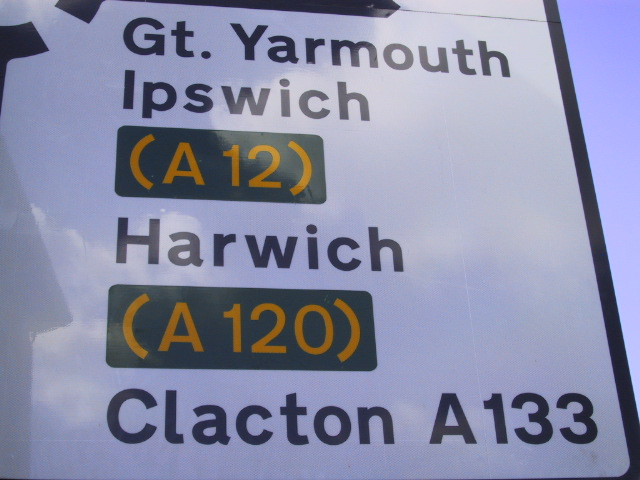}
\includegraphics[width=2.25cm,keepaspectratio]{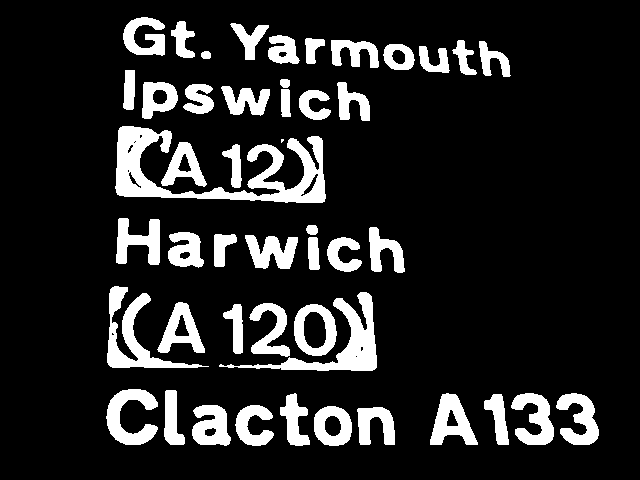}
\includegraphics[width=2.25cm,keepaspectratio]{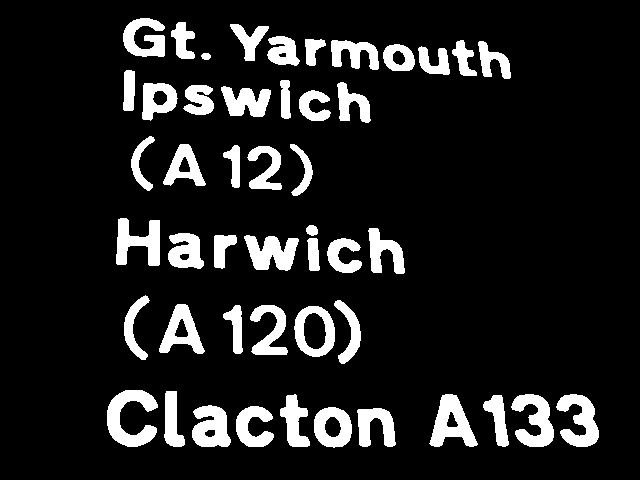}
\includegraphics[width=2.25cm,keepaspectratio]{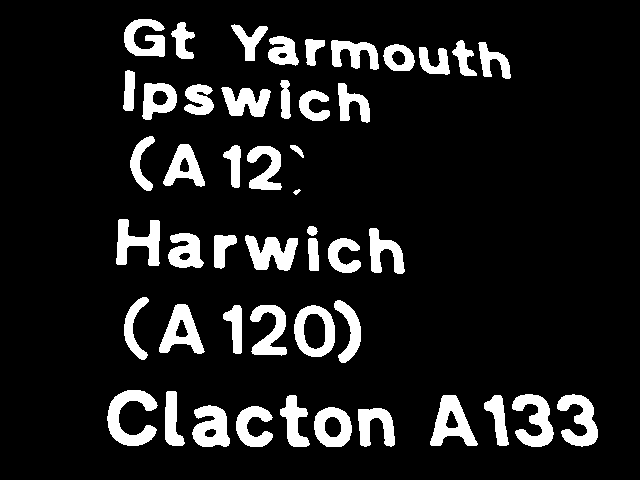}
\includegraphics[width=2.25cm,keepaspectratio]{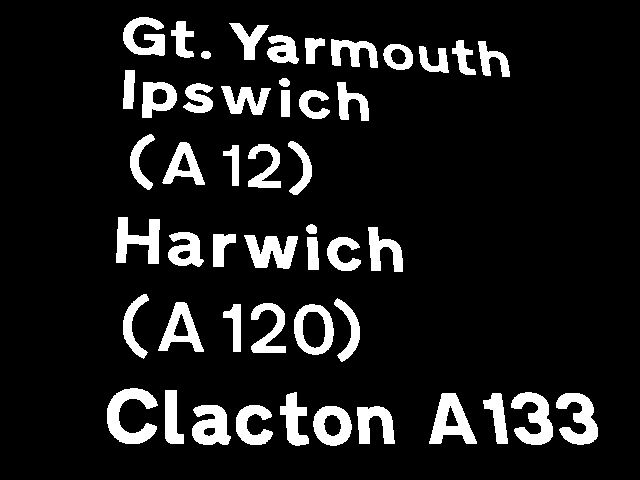}}
\vskip 0.08in
\centerline{
\includegraphics[width=2.25cm,keepaspectratio]{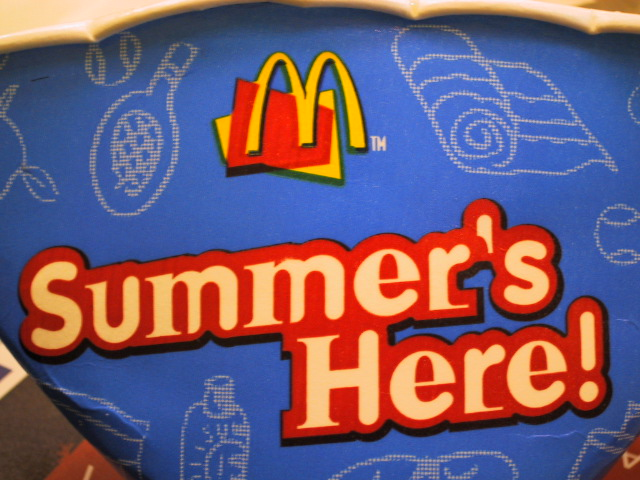}
\includegraphics[width=2.25cm,keepaspectratio]{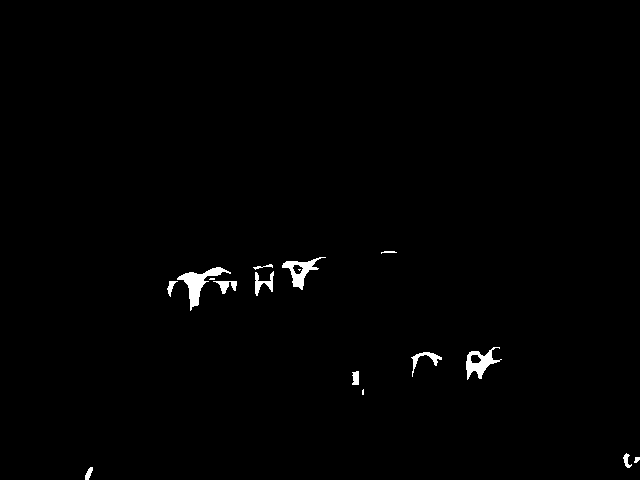}
\includegraphics[width=2.25cm,keepaspectratio]{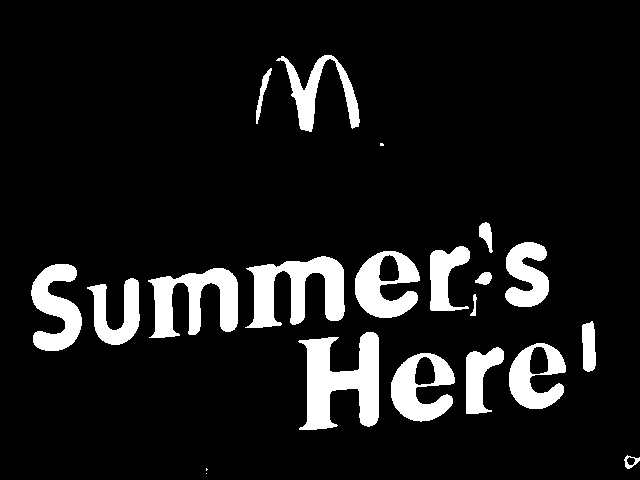}
\includegraphics[width=2.25cm,keepaspectratio]{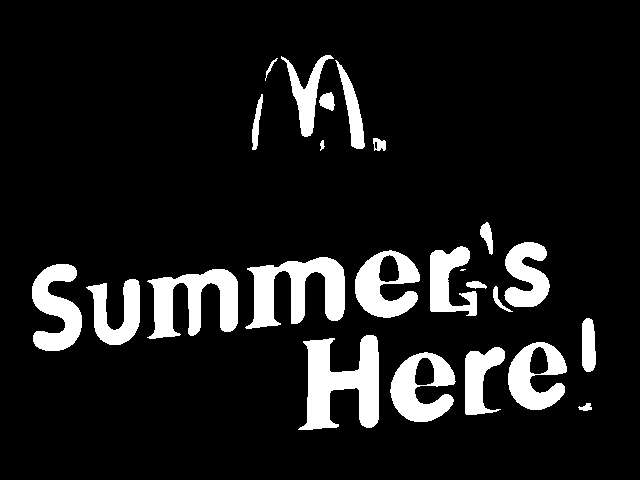}
\includegraphics[width=2.25cm,keepaspectratio]{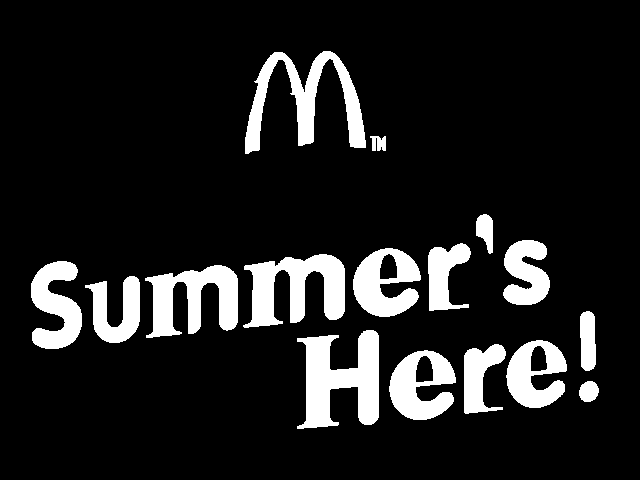}}
\vskip -0.05in
\centerline{
\subfloat[]{\includegraphics[width=2.25cm,keepaspectratio]{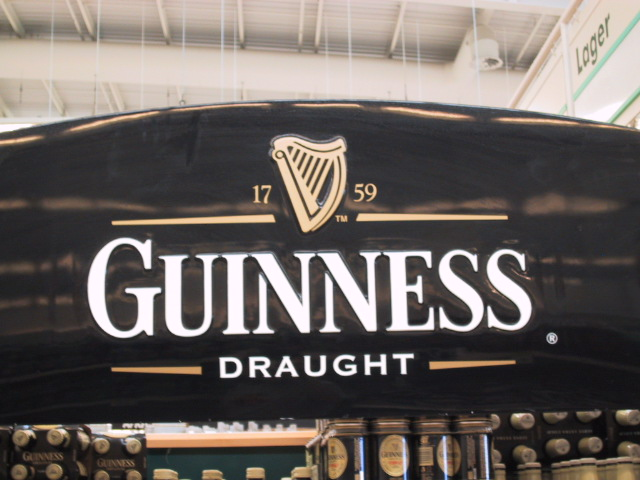}\hskip 0.045in}
\subfloat[]{\includegraphics[width=2.25cm,keepaspectratio]{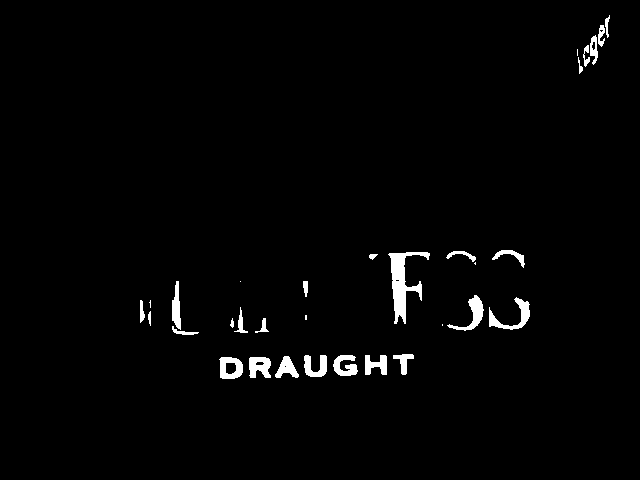}\hskip 0.045in}
\subfloat[]{\includegraphics[width=2.25cm,keepaspectratio]{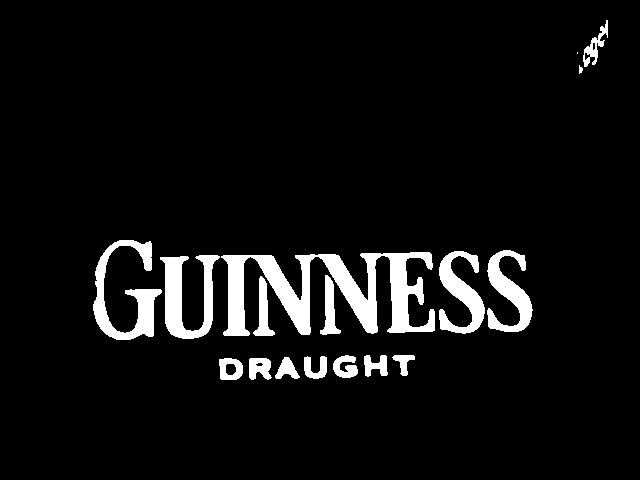}\hskip 0.045in}
\subfloat[]{\includegraphics[width=2.25cm,keepaspectratio]{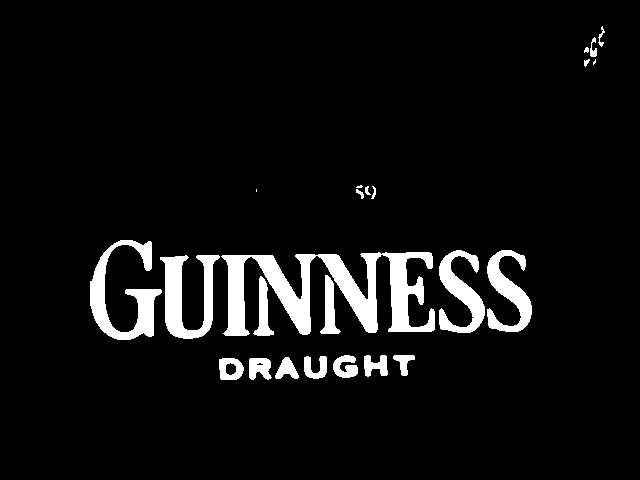}\hskip 0.045in}
\subfloat[]{\includegraphics[width=2.25cm,keepaspectratio]{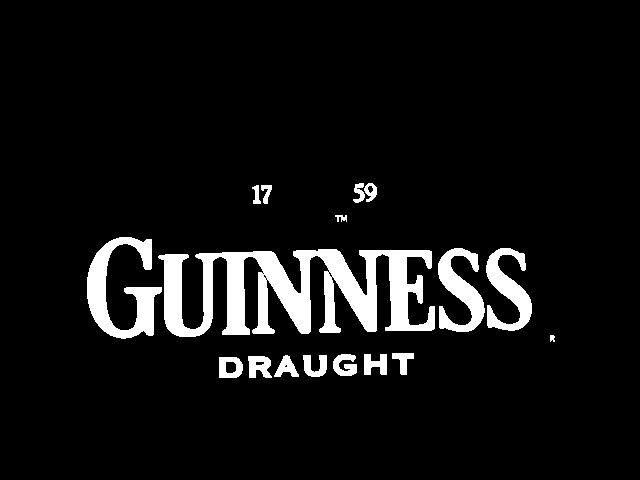}}}
\caption{Results on the ICDAR--2013 test set. In (a) the original image, in (b), (c) and (d) the segmentation obtained with Synth, Synth+COCO\_TS and COCO\_TS setups, respectively. The ground--truth supervision is reported in (e).}
\label{segmentation_results_ICDAR}
\end{center}
\begin{center}
\centerline{
\includegraphics[width=2.25cm,keepaspectratio]{img_segmentation/total_text/src/img2.png}
\includegraphics[width=2.25cm,keepaspectratio]{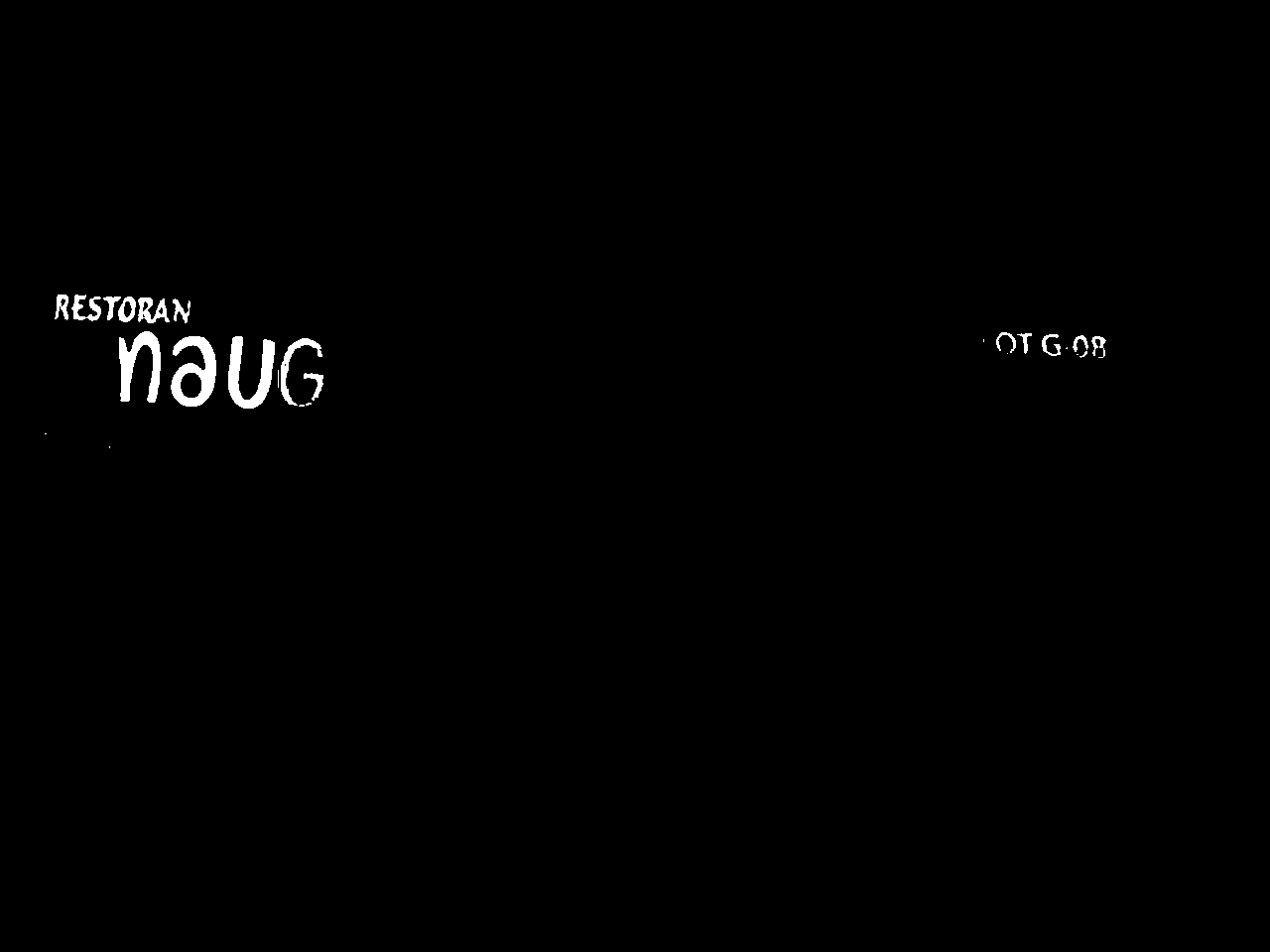}
\includegraphics[width=2.25cm,keepaspectratio]{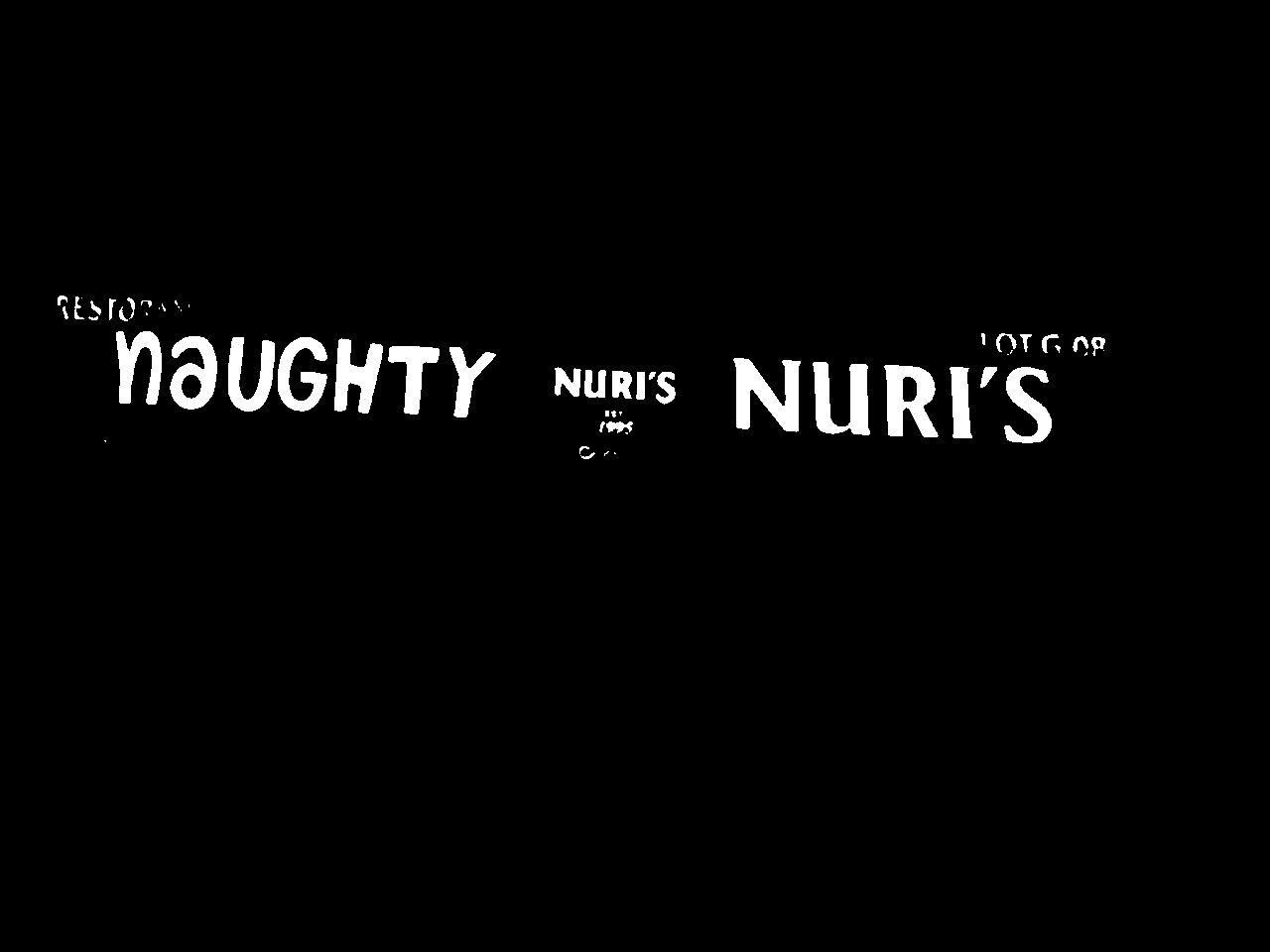}
\includegraphics[width=2.25cm,keepaspectratio]{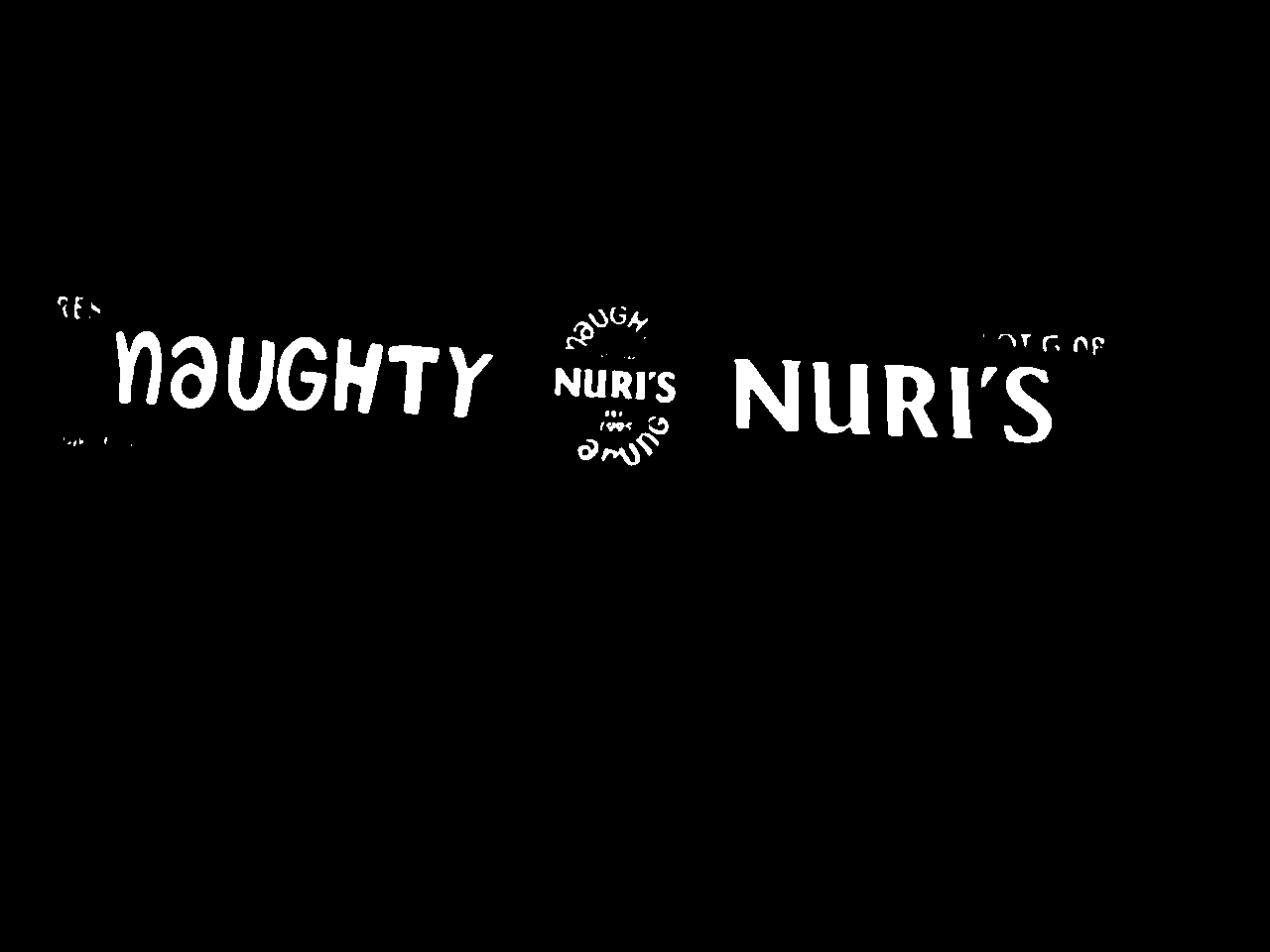}
\includegraphics[width=2.25cm,keepaspectratio]{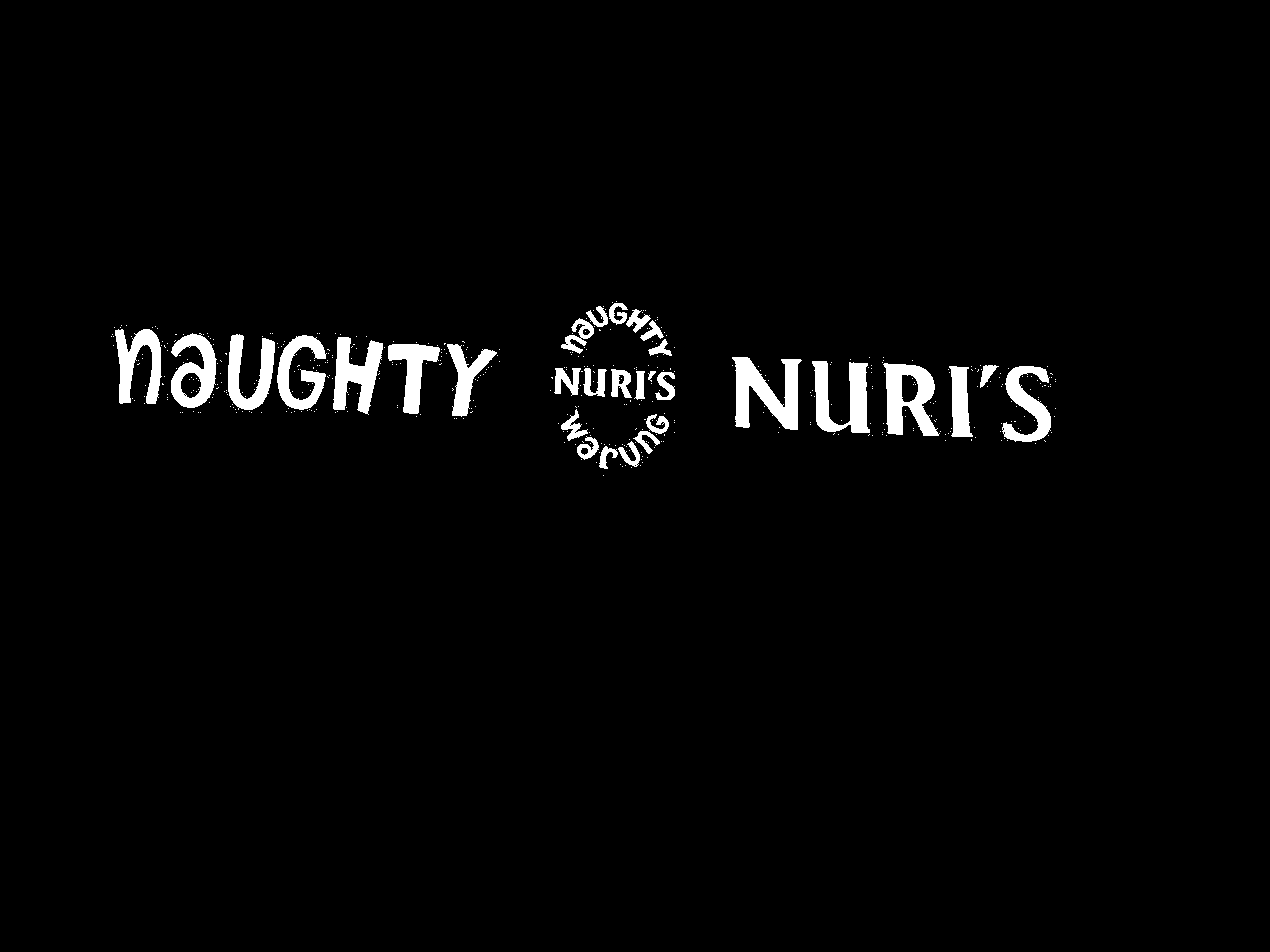}}
\vskip 0.08in
\centerline{
\includegraphics[width=2.25cm,keepaspectratio]{img_segmentation/total_text/src/img4.png}
\includegraphics[width=2.25cm,keepaspectratio]{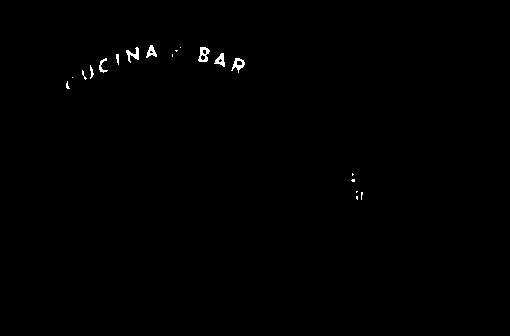}
\includegraphics[width=2.25cm,keepaspectratio]{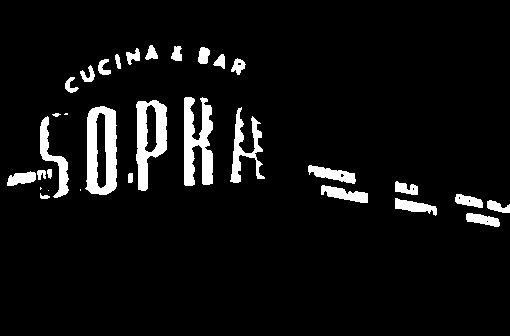}
\includegraphics[width=2.25cm,keepaspectratio]{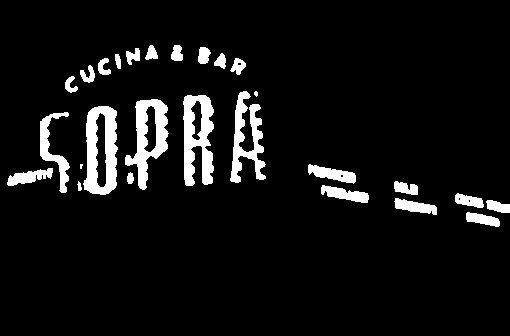}
\includegraphics[width=2.25cm,keepaspectratio]{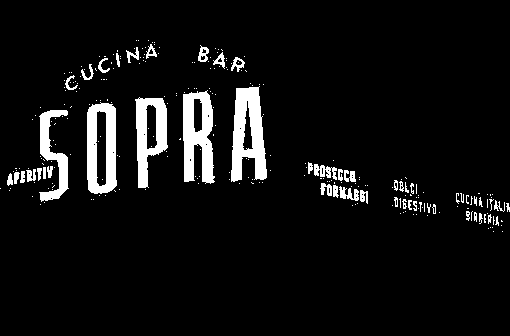}}
\vskip -0.05in
\centerline{
\subfloat[]{\includegraphics[width=2.25cm,keepaspectratio]{img_segmentation/total_text/src/img8.png}\hskip 0.045in}
\subfloat[]{\includegraphics[width=2.25cm,keepaspectratio]{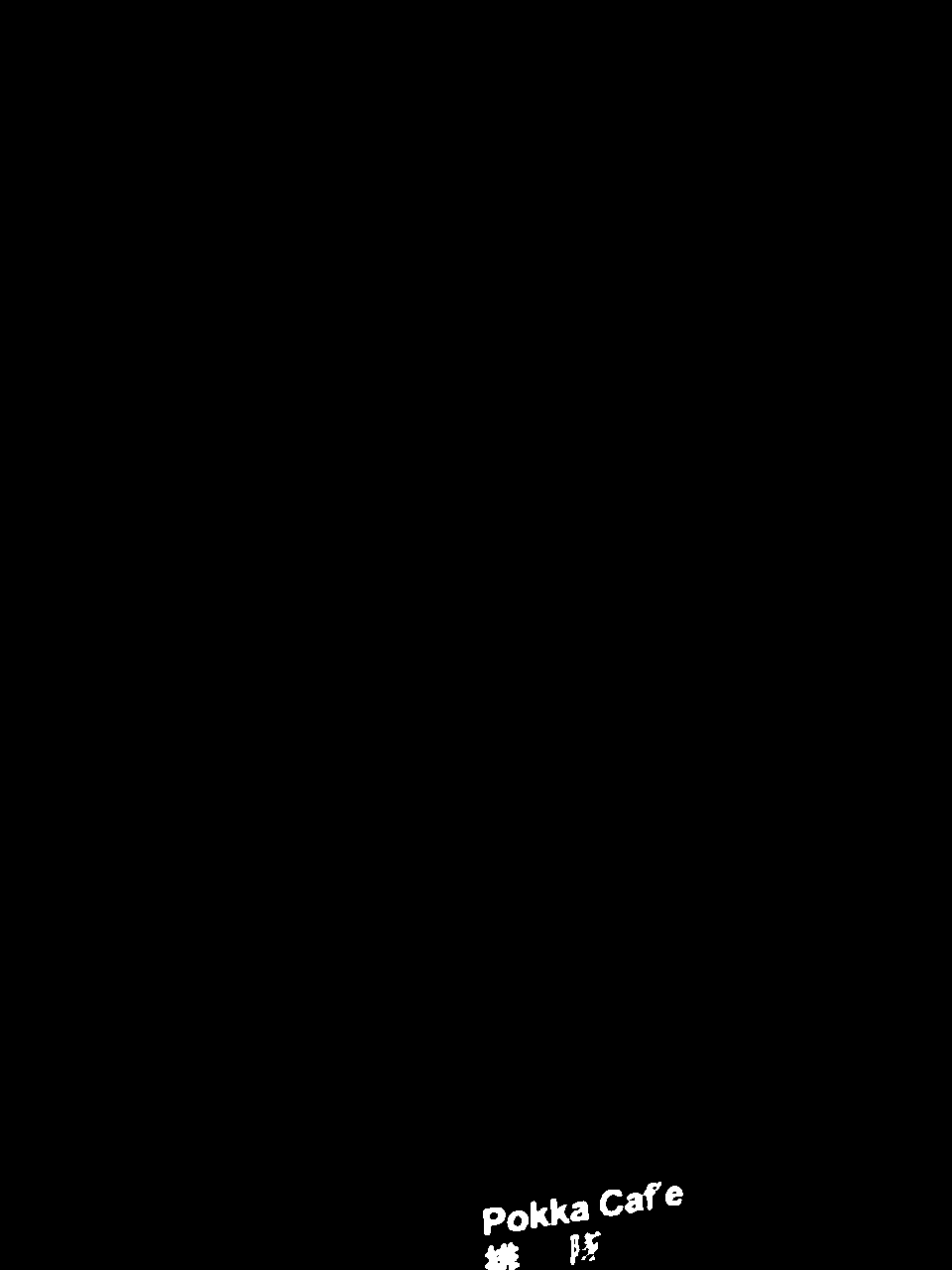}\hskip 0.045in}
\subfloat[]{\includegraphics[width=2.25cm,keepaspectratio]{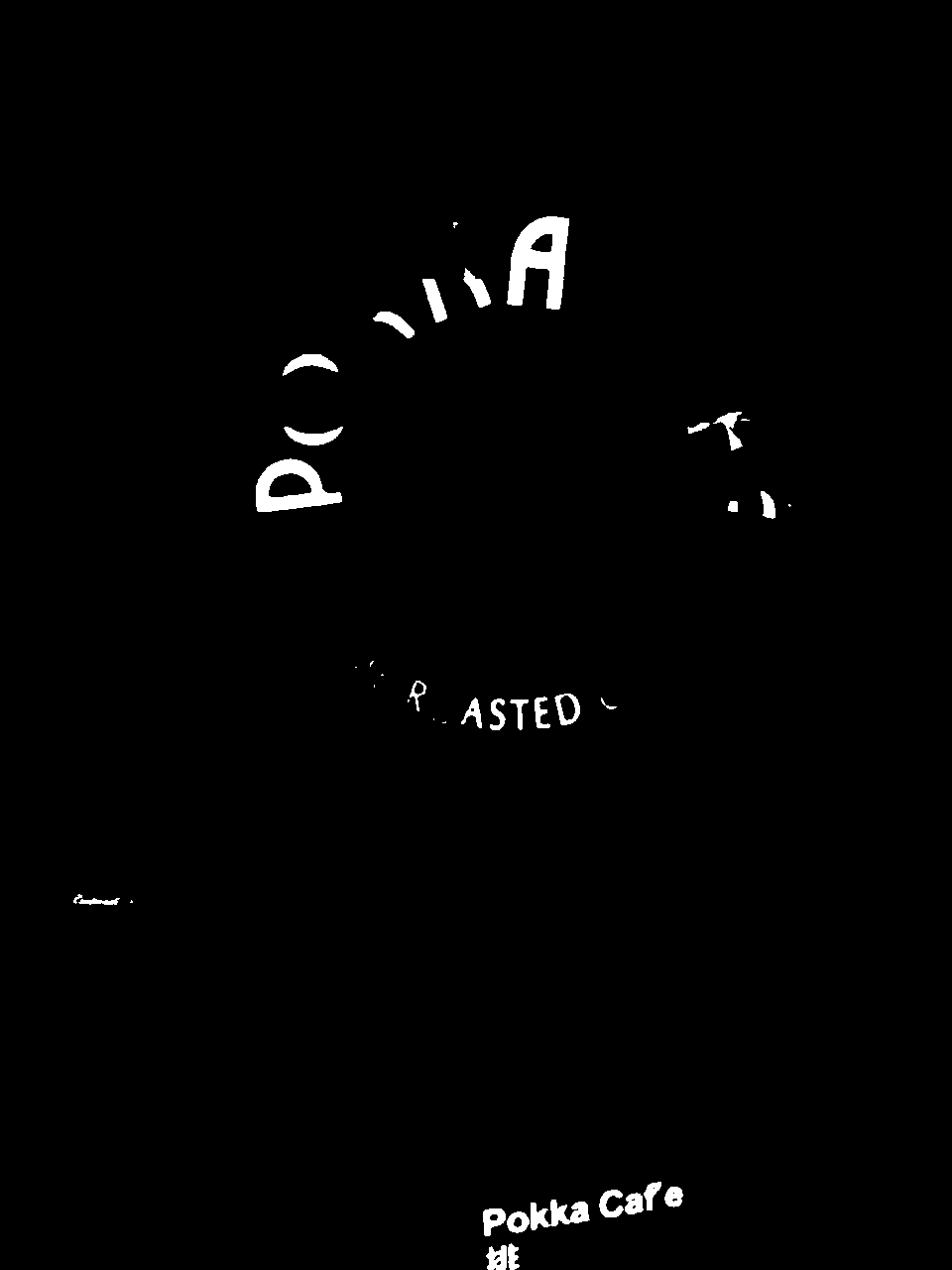}\hskip 0.045in}
\subfloat[]{\includegraphics[width=2.25cm,keepaspectratio]{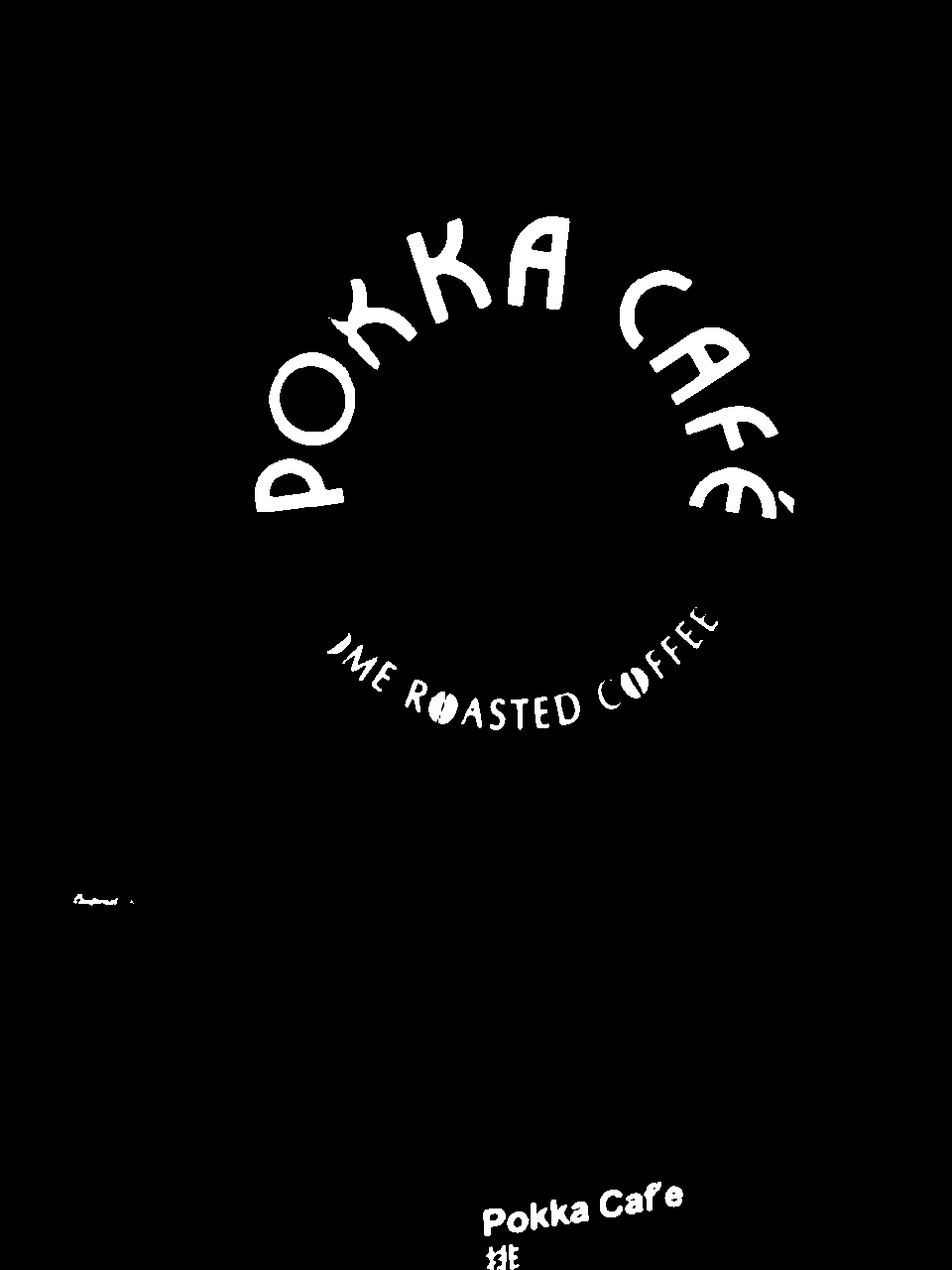}\hskip 0.045in}
\subfloat[]{\includegraphics[width=2.25cm,keepaspectratio]{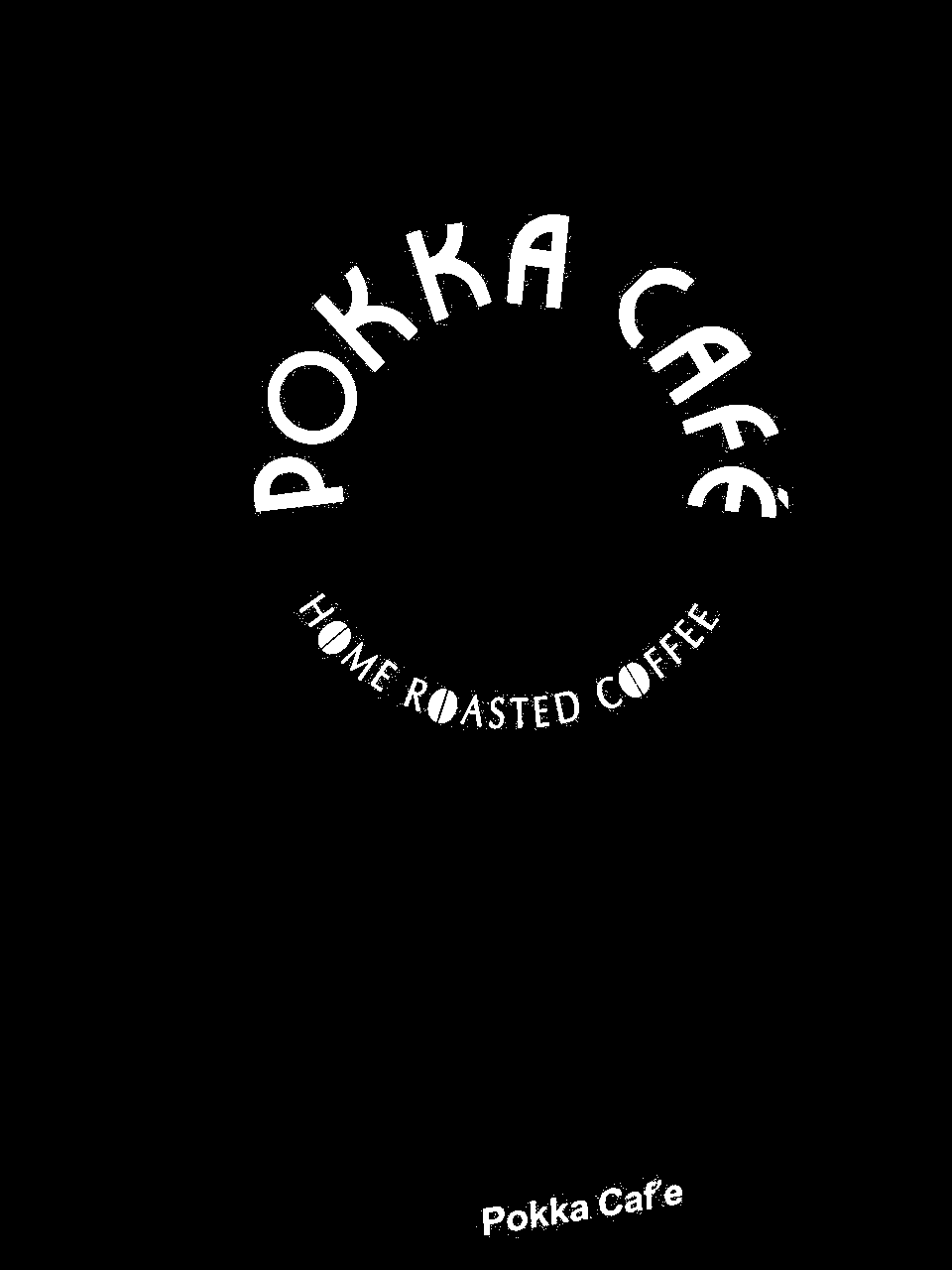}}}
\caption{Results on the Total--Text test set. In (a) the original image, in (b), (c) and (d) the segmentation obtained with Synth, Synth+COCO\_TS and COCO\_TS setup, respectively. The ground--truth supervision is reported in (e).}
\label{segmentation_results_TotalText}
\end{center}
\end{figure}

Due to the inherent difficulties in collecting large sets of pixel--level supervised images, only few public datasets are available for scene text segmentation. To face this problem, in \cite{tang2017scene}, synthetic data generation has been employed. Nevertheless, due to the domain--shift, there is no guarantee that a network trained on synthetic data would generalize well also to real images. 
The COCO\_TS dataset actually contains real images and, therefore, we expect that, when used for network training, the domain--shift can be reduced. To test this hypothesis, the PSPNet is used for scene text segmentation and evaluated on the ICDAR--2013 and Total--Text test sets, that provides pixel--level annotations. In particular, the following experimental setups have been compared:
 \begin{itemize} 
\item \textbf{Synth:} The training relies only on the synthetically generated images;
\item \textbf{Synth + COCO\_TS:} The network is pre--trained on the synthetic dataset and fine--tuned on the COCO\_TS images;
\item \textbf{COCO\_TS:} The network is trained only on the COCO\_TS dataset. 
 \end{itemize} 
\noindent The influence of fine--tuning on the ICDAR--2013 and Total--Text datasets was also evaluated. The results, measured using the pixel--level precision, recall and F1 score, are reported in Table \ref{ICDAR2013_Results} and Table \ref{TotalText_Results}, respectively.
\begin{table*}[ht]
\label{icdar_results}
\begin{center}
\begin{small}
\subfloat[\small{Results on the ICDAR--2013 test set} \label{ICDAR2013_Results}]{%
\begin{tabular}{|lcccc|}
\hline
 & Precision & Recall & F1 Score & \\
\hline
Synth & 73.19\%& 55.67\% & 63.23\% & --\\
Synth + COCO\_TS & \textbf{77.80\%}& \textbf{70.14\%} & \textbf{73.77\%} & \textbf{+10.54\%} \\
COCO\_TS & 78.86\% & 68.66\% & 73.40\% & +10.17\%\\
\hline
Synth + ICDAR--2013 & 81.12\%& 78.33\% & 79.70\% & --\\
Synth + COCO\_TS + ICDAR--2013  & 80.08\%& 79.53\% & 80.15\% & +0.45\%\\
COCO\_TS + ICDAR--2013  & \textbf{81.68\%} & \textbf{79.16\%} & \textbf{80.40\%} & \textbf{+0.70\%}\\
\hline
\end{tabular}}
\end{small}
\begin{small}
\subfloat[\small{Results on the Total--Text test set} \label{TotalText_Results}]{%
\begin{tabular}{|lcccc|}
\hline
 & Precision & Recall & F1 Score & \\
\hline
Synth & 55.76\%& 22.87\% & 32.43\% & --\\
Synth + COCO\_TS & 72.71\%& 54.49\% & 62.29\% & +29.86\% \\
COCO\_TS & \textbf{72.83\%} & \textbf{56.81\%} & \textbf{63.83\%} & \textbf{+31.40\%}\\
\hline
Synth + Total Text & 84.97\%& 65.52\% & 73.98\% & --\\
Synth + COCO\_TS + Total Text  & 84.65\%& 66.93\% & 74.75\% & +0.77\%\\
COCO\_TS + Total Text  & \textbf{84.31\%} & \textbf{68.03\%} & \textbf{75.30\%} & \textbf{+1.32\%}\\
\hline
\end{tabular}}
\end{small}
\end{center}
\caption{Scene text segmentation performances using synthetic data and/or the proposed COCO\_TS dataset. The notation "$+$ Dataset" means that a fine--tune procedure has been carried out on "Dataset". The last column reports the relative increment, with and without fine--tuning, compared to the use of synthetic data only.}
\end{table*}
\noindent It is worth noting that training the network using the COCO\_TS dataset is more effective than using synthetic images. Specifically, employing the proposed dataset, the F1 Score is improved of 10.17\% and 31.40\% on ICDAR--2013 and Total--Text, respectively. 
These results are quite surprising and prove that the proposed dataset substantially increases the network performance, reducing the domain--shift from synthetic to real images. If the network is fine--tuned on ICDAR--2013 or Total--Text, the relative difference between the use of synthetic images and the COCO\_TS dataset is reduced, but still remains significant. Specifically, the F1 Score is improved by 0.70\% on ICDAR--2013 and 1.32\% on Total--Text. 
\noindent Furthermore, it can be observed that using only COCO\_TS provides comparable results than training the network with both the synthetic and the proposed dataset. Therefore, the two datasets are not complementary and, in fact, the proposed COCO\_TS is a valid alternative to synthetic data generation for scene text segmentation. Indeed, the use of real images increases the sample efficiency, allowing to substantially reduce the number of samples needed for training. In particular, the COCO\_TS dataset contains 14690 samples that are less than 1/50 of the synthetic dataset cardinality.  
Some qualitative output results of the scene text segmentation network are shown in Figure \ref{segmentation_results_ICDAR} and Figure \ref{segmentation_results_TotalText}.

\section{Conclusions} \label{Conclusions}
In this paper, a weakly supervised learning approach has been used to generate pixel--level supervisions for scene text segmentation. Exploiting the proposed approach, the COCO\_TS dataset, which contains the segmentation ground--truth for a subset of the COCO--Text dataset, has been automatically generated. Unlike previous approaches based on synthetic images, a convolutional neural network is trained on real images from the COCO\_TS dataset for scene text segmentation, showing a very significant improvement in the generalization on both the ICDAR--2013 and Total--Text datasets, although with only a fraction of the samples. To foster further research on scene text segmentation, the COCO\_TS dataset has been released.
Interestingly, our procedure for pixel--level supervision generation from bounding--box annotations is general and not limited to the COCO--Text dataset. It is a matter of future work to employ the same method to extract pixel--level supervisions for different text localization problems (f.i., on multilingual scene text datasets, such as MLT \cite{MLT}).

%
%
%

\begin{thebibliography}{10}
\providecommand{\url}[1]{\texttt{#1}}
\providecommand{\urlprefix}{URL }
\providecommand{\doi}[1]{https://doi.org/#1}

\bibitem{icdar2013}
Karatzas, D., Shafait, F., Uchida, S., Iwamura, M., i~Bigorda, L.G., Mestre,
  S.R., Mas, J., Mota, D.F., Almazan, J.A., De~Las~Heras, L.P.: Icdar 2013
  robust reading competition. In: Document Analysis and Recognition (ICDAR),
  2013 12th International Conference on. pp. 1484--1493. IEEE (2013)

\bibitem{totalText}
Ch’ng, C.K., Chan, C.S.: Total-text: A comprehensive dataset for scene text
  detection and recognition. In: 14th IAPR International Conference on Document
  Analysis and Recognition {ICDAR}. pp. 935--942 (2017).
  \doi{10.1109/ICDAR.2017.157}
  
\bibitem{tang2017scene}
Tang, Y., Wu, X.: Scene text detection and segmentation based on cascaded
  convolution neural networks. IEEE Transactions on Image Processing
  \textbf{26}(3),  1509--1520 (2017)
  
\bibitem{synthText}
Gupta, A., Vedaldi, A., Zisserman, A.: Synthetic data for text localisation in
  natural images. In: IEEE Conference on Computer Vision and Pattern
  Recognition (2016)

\bibitem{cocotext}
Veit, A., Matera, T., Neumann, L., Matas, J., Belongie, S.: Coco-text: Dataset
  and benchmark for text detection and recognition in natural images. In: arXiv
  preprint arXiv:1601.07140 (2016),
  \url{http://vision.cornell.edu/se3/wp-content/uploads/2016/01/1601.07140v1.pdf}
  
\bibitem{icdar2015}
Karatzas, D., Gomez-Bigorda, L., Nicolaou, A., Ghosh, S., Bagdanov, A.,
  Iwamura, M., Matas, J., Neumann, L., Chandrasekhar, V.R., Lu, S., et~al.:
  Icdar 2015 competition on robust reading. In: Document Analysis and
  Recognition (ICDAR), 2015 13th International Conference on. pp. 1156--1160.
  IEEE (2015)
  
\bibitem{MLT}
RRC-MLT: Mlt: Competition on multi-lingual scene text detection and script
  identification. http://rrc.cvc.uab.es/?ch=8\&com=introduction, accessed:
  2019-01-01

\bibitem{bbox}
Bonechi, S., Andreini, P., Bianchini, M., Scarselli, F.: Generating bounding
  box supervision for semantic segmentation with deep learning. In: IAPR
  Workshop on Artificial Neural Networks in Pattern Recognition. pp. 190--200.
  Springer (2018)

\bibitem{SYNTHIA}
Ros, G., Sellart, L., Materzynska, J., Vazquez, D., Lopez, A.M.: The synthia
  dataset: A large collection of synthetic images for semantic segmentation of
  urban scenes. In: Proceedings of the IEEE conference on computer vision and
  pattern recognition. pp. 3234--3243 (2016)

\bibitem{indoor}
Handa, A., Patraucean, V., Badrinarayanan, V., Stent, S., Cipolla, R.:
  Synthcam3d: Semantic understanding with synthetic indoor scenes. arXiv
  preprint arXiv:1505.00171  (2015)

\bibitem{synth_colony}
Andreini, P., Bonechi, S., Bianchini, M., Mecocci, A., Scarselli, F.: A deep
  learning approach to bacterial colony segmentation. In: International
  Conference on Artificial Neural Networks. pp. 522--533. Springer (2018)

\bibitem{jaderberg_text}
Jaderberg, M., Simonyan, K., Vedaldi, A., Zisserman, A.: Reading text in the
  wild with convolutional neural networks. International Journal of Computer
  Vision  \textbf{116}(1),  1--20 (2016)
  
\bibitem{BoxSup}
Dai, J., He, K., Sun, J.: Boxsup: Exploiting bounding boxes to supervise
  convolutional networks for semantic segmentation. In: Proceedings of the IEEE
  International Conference on Computer Vision. pp. 1635--1643 (2015)

\bibitem{EM_weak}
Papandreou, G., Chen, L.C., Murphy, K., Yuille, A.L.: Weakly-and
  semi-supervised learning of a dcnn for semantic image segmentation. arXiv
  preprint arXiv:1502.02734  (2015)  
  
\bibitem{bbox_khoreva}
Khoreva, A., Benenson, R., Hosang, J., Hein, M., Schiele, B.: Simple does it:
  Weakly supervised instance and semantic segmentation. In: Proceedings of the
  IEEE conference on computer vision and pattern recognition. pp. 876--885
  (2017)

\bibitem{FCN}
Long, J., Shelhamer, E., Darrell, T.: Fully convolutional networks for semantic
  segmentation. In: Proceedings of the IEEE conference on computer vision and
  pattern recognition. pp. 3431--3440 (2015)

\bibitem{atrous}
Papandreou, G., Kokkinos, I., Savalle, P.A.: Untangling local and global
  deformations in deep convolutional networks for image classification and
  sliding window detection. arXiv preprint arXiv:1412.0296  (2014)

\bibitem{PSP}
Zhao, H., Shi, J., Qi, X., Wang, X., Jia, J.: Pyramid scene parsing network.
  2017 IEEE Conference on Computer Vision and Pattern Recognition (CVPR) pp.
  6230--6239 (2017)

\bibitem{deeplab}
Chen, L.C., Papandreou, G., Kokkinos, I., Murphy, K., Yuille, A.L.: Deeplab:
  Semantic image segmentation with deep convolutional nets, atrous convolution,
  and fully connected crfs. IEEE transactions on pattern analysis and machine
  intelligence  \textbf{40}(4),  834--848 (2017)
  
\bibitem{otsu}
Otsu, N.: A threshold selection method from gray-level histograms. IEEE
  Transactions on Systems, Man and Cybernetics  \textbf{9}(1),  62--66 (1979)
  
\bibitem{binarization}
Su, B., Lu, S., Tan, C.L.: Binarization of historical document images using the
  local maximum and minimum. In: Proceedings of the 9th IAPR International
  Workshop on Document Analysis Systems. pp. 159--166. ACM (2010)
  
\bibitem{laplacian}
Howe, N.R.: A laplacian energy for document binarization. In: Document Analysis
  and Recognition (ICDAR), 2011 International Conference on. pp. 6--10. IEEE
  (2011)

\bibitem{bai2014seed}
Bai, B., Yin, F., Liu, C.L.: A seed-based segmentation method for scene text
  extraction. In: Document Analysis Systems (DAS), 2014 11th IAPR International
  Workshop on. pp. 262--266. IEEE (2014)

\bibitem{mishra2011mrf}
Mishra, A., Alahari, K., Jawahar, C.: An mrf model for binarization of natural
  scene text. In: ICDAR-International Conference on Document Analysis and
  Recognition. IEEE (2011)

\bibitem{tian2014scene}
Tian, S., Lu, S., Su, B., Tan, C.L.: Scene text segmentation with multi-level
  maximally stable extremal regions. In: 2014 22nd International Conference on
  Pattern Recognition (ICPR). pp. 2703--2708. IEEE (2014)

\bibitem{coco}
Lin, T.Y., Maire, M., Belongie, S., Hays, J., Perona, P., Ramanan, D.,
  Doll{\'a}r, P., Zitnick, C.L.: Microsoft coco: Common objects in context. In:
  European conference on computer vision. pp. 740--755. Springer (2014)

\bibitem{icdar2011}
Shahab, A., Shafait, F., Dengel, A.: Icdar 2011 robust reading competition
  challenge 2: Reading text in scene images. In: Document Analysis and
  Recognition (ICDAR), 2011 International Conference on. pp. 1491--1496. IEEE
  (2011)

\bibitem{adam}
Kingma, D., Ba, J.: Dp kingma and j. ba, adam: A method for stochastic
  optimization, arxiv: 1412.6980. Adam: A Method for Stochastic Optimization


\end{thebibliography}

\end{document}